# De-noising, Stabilizing and Completing 3D Reconstructions On-the-go using Plane Priors

Maksym Dzitsiuk[1,2], Jürgen Sturm[2], Robert Maier[1], Lingni Ma[1] and Daniel Cremers[1]

*Abstract* — Creating 3D maps on robots and other mobile devices has become a reality in recent years. Online 3D reconstruction enables many exciting applications in robotics and AR/VR gaming. However, the reconstructions are noisy and generally incomplete. Moreover, during online reconstruction, the surface changes with every newly integrated depth image which poses a significant challenge for physics engines and path planning algorithms. This paper presents a novel, fast and robust method for obtaining and using information about planar surfaces, such as walls, floors, and ceilings as a stage in 3D reconstruction based on Signed Distance Fields (SDFs). Our algorithm recovers clean and accurate surfaces, reduces the movement of individual mesh vertices caused by noise during online reconstruction and fills in the occluded and unobserved regions. We implemented and evaluated two different strategies to generate plane candidates and two strategies for merging them. Our implementation is optimized to run in real-time on mobile devices such as the Tango tablet. In an extensive set of experiments, we validated that our approach works well in a large number of natural environments despite the presence of significant amount of occlusion, clutter and noise, which occur frequently. We further show that plane fitting enables in many cases a meaningful semantic segmentation of real-world scenes.

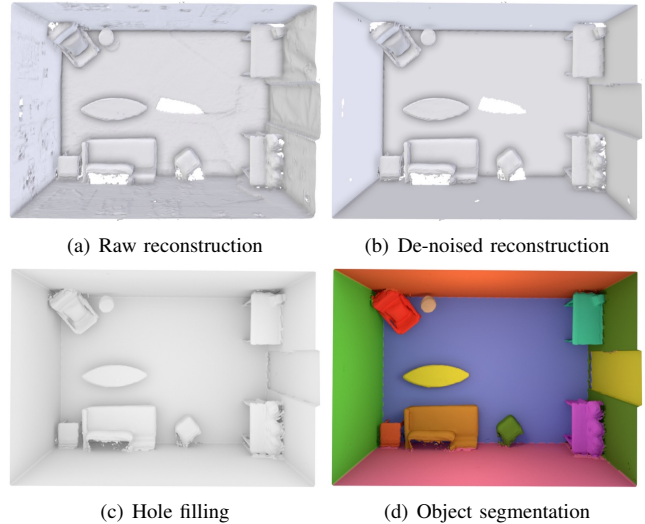

Fig. 1. Real-world scans generally suffer from noise and holes in the reconstructed models (top left). In man-made environments, the noise can be significantly reduced by applying a plane prior (top right). Furthermore, by extending the detected planes into unobserved regions, holes can be filled automatically (bottom left). This also enables scene segmentation and object detection (bottom right).

## I. INTRODUCTION

In this paper we investigate the problem of real-time, on-device 3D reconstruction. While existing 3D reconstruction algorithms [1]–[6] are mostly targeted at offline use of the resulting 3D models, many applications in robotics and AR gaming require that valid and dependable models exist already during model acquisition, e.g., because the robot or player actively explores the environment while the 3D model is being acquired.

This leads to several interesting problems that haven't received much attention so far and that we tackle in this paper: First, the 3D model changes with every new observation, which leads to a wobbling surface and changing topology (see Fig. 3 for an analysis), which poses a problem for path planning algorithms and physics simulations. For example, virtual balls thrown into the mesh of a currently reconstructed room will continuously roll around as the reconstruction gets refined. A tower of virtual objects falls without user intervention for the same reason (Fig. 2). Second, the 3D reconstruction typically contains holes and is generally incomplete, even after scanning has been finalized. Similarly, this causes problems for robot navigation and simulated avatars that cannot reason about space outside of the already scanned area. Third, the reconstructions are generally noisy (especially in the beginning) which is (at least) visually unpleasant if not problematic for traversability analysis.

Our goal is thus to de-noise, stabilize and complete 3D scans during model acquisition. While this is a hard problem in the general case, it becomes tractable in indoor environments where we may expect many planar surfaces. Incorporating plane priors into 3D indoor reconstruction scenarios yields cleaner, more realistic and more complete 3D models of rooms or house-scale environments by improving the geometric reconstruction of walls, floors and other planar surfaces. Figure 1 illustrates our approach. It is clearly visible that the noise is significantly reduced on planar surfaces resulting in a more accurate and visually more appealing geometric 3D reconstruction. Furthermore, we are able to close various holes in the floor and the walls. Finally, objects can be segmented and semantic information can be attached to the reconstructed scene.

To achieve high-quality reconstructions from commodity RGB-D sensors, many approaches fuse noisy raw depth maps in memory efficient data structures such as signed distance fields (SDF) [7]. These implicit surface representations allow for highly accurate geometric recontructions and provide the required scalability and memory efficiency. In this paper, we introduce an efficient method to apply a plane prior during SDF-based 3D reconstruction. The accompanying video can be found at https://youtu.be/8OxwiRpmzn4. The main contributions of this paper are:

- We propose an algorithm for real-time 3D reconstruction on mobile devices that incorporates priors for modelling planar surfaces of indoor environments.
- We introduce a novel way of estimating planes by fitting them to Signed Distance Fields. The presented method is based on robust least squares and is faster than RANSAC-based methods. The detected local plane candidates are merged to find globally consistent planes.
- We apply our approach to significantly reduce noise on flat surfaces by correcting SDF values using detected planes. Further, we demonstrate that the method can be used to fill holes in incomplete reconstructed 3D models. Lastly, we use plane priors to directly obtain object-level and semantic segmentation (e.g. walls) of indoor environments.

[1]Maksym Dzitsiuk, Robert Maier, Lingni Ma and Daniel Cremers are with the Computer Vision Group, Department of Computer Science, Technical University of Munich {dzitsiuk,maierr,lingni,cremers}@in.tum.de.
[2]Maksym Dzitsiuk and Jürgen Sturm are with Google {dzitsiuk,jsturm}@google.com.

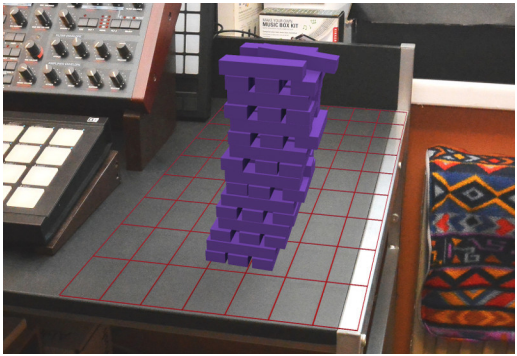

Fig. 2. Continuous 3D reconstruction is problematic for physics engines. This image shows a virtual Jenga tower standing on a planar surface. Due to sensor noise during reconstruction, the tower falls over without applying a plane prior.

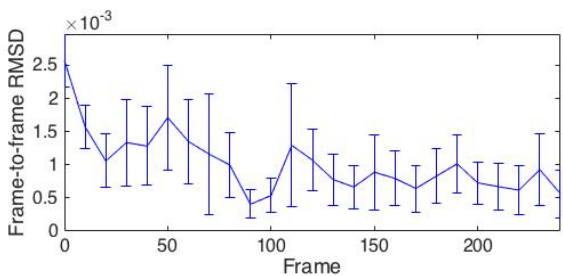

Fig. 3. During online 3D reconstruction, the surface is constantly moving due to sensor noise. The plot shows the root mean square difference (RMSD) between vertices of the reconstructed mesh over time. The examined data includes static observation of floor over 250 frames, executed 10 times.

## II. RELATED WORK

In this section, we first discuss the most related works of state-of-the-art RGB-D based 3D reconstruction. Second, we discuss the use of plane priors in 2.5/3D reconstruction.

**3D reconstruction with signed distance fields:** KinectFusion by Newcombe et al. [1] was the first system to reconstruct highly accurate 3D models of persons, objects and workspaces of limited size. Extensions to KinectFusion focus on various aspects of 3D modeling. Bylow et al. [8] perform direct camera tracking against the SDF, while Niessner et al. [3] enable large-scale 3D mapping using an efficient voxel hashing scheme. Other approaches tackle the problem of large-scale 3D reconstruction by direct frame-to-(key)frame tracking with continuous pose graph optimization [2] and subsequent data fusion in an Octree-based SDF representation [9] or by achieving global consistency using submap-based bundle adjustment [10]. Recently, the BundleFusion system by Dai et al. [6] extends [3] to allow for large-scale reconstruction of previously unseen geometric accuracy. However, none of these frameworks exploits plane priors for tracking or improving the reconstruction quality. As these systems usually extensively rely on the use of GPUs or powerful CPUs, real-time 3D reconstruction approaches for the application on mobile devices have been designed specifically. Kähler et al. [5] modified the voxel hashing framework [3] to efficiently work in real-time on a mobile NVIDIA Shield Tablet with a built-in GPU. Finally, our work is based on CHISEL [4] which enables dense housescale 3D reconstructions in real-time on a Tango device. The system does not rely on a GPU and combines visual-inertial odometry for localization with an efficient spatially hashed SDF volume for mapping. Our approach uses this implicit surface representation and combines it with plane priors in order to improve reconstructions in various ways.

**Incorporating geometric priors during 3D reconstruction:** Plane segmentation is a useful preprocessing for many indoor environment mapping and tracking algorithms. Given 3D point clouds, one commonly used plane segmentation algorithm is based on RANdom SAmple Consensus (RANSAC) [11]. RANSAC-based algorithms find the point cluster that best fits a plane model, but they also produce false detections in many situations because of not considering underlying connectivity of points. To tackle this drawback, many solutions have been developed based on region-growing [12]–[15]. When using RGB-D cameras, direct plane detection on depth maps become possible, since dense depth maps embed organized point clouds. Holz et al. [16] proposed a fast plane detection which performs fast normal estimation using integral images and subsequently clusters points into planar regions according to the normals. Feng et al. [17] developed an agglomerative hierarchical clustering (AHC) algorithm which finds blockwise planar regions with least squares plane fitting and then merges the local regions to form complete plane segments. Jiang et al. [18] fit cuboids to depth images by first dividing the image into planar superpixels and then using RANSAC to generate suitable cuboid candidates. Silberman et al. [19] use a probabilistic model to complete the boundaries of occluded objects by detecting the dominant planes from partial voxel based reconstruction. Most directly related to our approach is the work by Zhang et al. [15], which segments incoming depth images and then similarly maintains a set of planes and propagates this information to the SDF.

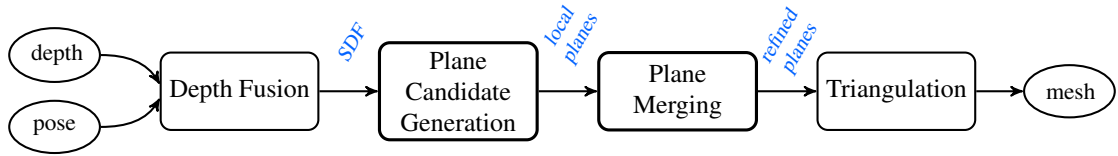

Fig. 4. **Pipeline overview.** Our main approach adds two steps (bold nodes) to the 3D reconstruction pipeline. Local plane candidate generation happens directly on the SDF grid by using an adapted least squares fitting approach. After local planes are estimated, they are clustered into fewer refined planes that represent real objects in the scene. The refined plane models can be used for any purpose, e.g. during triangulation to reduce the noise on planar surfaces.

Despite all proposed plane segmentation algorithms, none of the solutions have been developed for implicit surface representations such as SDFs. Additionally to evaluating RANSAC on the reconstructed mesh, we have developed a novel approach of fitting planes directly to SDF grids, without the need for an explicit surface representation. As evaluation shows, this approach significantly reduces the runtime of the fitting process and does not require additional storage, making it applicable on mobile devices and robots.

## III. 3D RECONSTRUCTION PIPELINE

The Tango visual inertial odometry (VIO) algorithm uses a fish-eye camera and an IMU to track its 3D pose in real-time. In post-processing, the camera poses can be refined using bundle adjustment to further reduce drift. We implemented our approach on top of CHISEL [4], which is the 3D reconstruction pipeline in Project Tango. At its heart, CHISEL uses a two level data structure to store the SDF: The first level contains blocks of voxels, which are called *volumes* throughout this paper. Volumes are spatially hashed into a dynamic 3D map based on their integer coordinates. They are allocated and deallocated dynamically to store only the areas where measurements are observed. Each volume consists of a fixed 3D grid of voxels, which are stored sequentially in memory. Each voxel contains an estimate of the SDF and an associated weight. When incorporating a new depth image, volumes get marked for re-meshing. After the update is complete, we extract a mesh of every volume that has been marked using Marching Cubes.

Typically, every voxel has a size of 3cm and every volume consists of $16^3$ voxels (corresponding to a cube of 48cm side length). The depth camera has a resolution of 160×120 pixels and operates at a framerate of 5 fps. All of our algorithms have been designed to run in real-time on a Tango device, except if noted otherwise.

## IV. PLANE SEGMENTATION

Our segmentation algorithm consists of two stages (Fig. 4): First, we detect planar surfaces locally in each volume and estimate plane candidate parameters (*local fitting, candidate generation*). Second, we use these plane candidates to estimate parameters of larger, multi-volume plane models (*clustering, merging*). These planes, that we also use to generate final mesh, are called *refined* throughout the paper.

As described in Sec. III, our surface representation is stored as a spatially hashed grid of volumes that contain SDF voxels. We use the same data structure to store information about detected and refined planes for every volume.

### A. Notation

We denote a 3D position with $\mathbf{x} = (x\ y\ z)^\top \in \mathbb{R}^3$. If needed, we denote with $\bar{\mathbf{x}} = (x\ y\ z\ 1)^\top \in \mathbb{R}^4$ the augmented vector. A 3D plane in Hessian normal form is denoted as $\mathbf{p} = (n_1\ n_2\ n_3\ d)^\top \in \mathbb{R}^4$ where $\mathbf{n} = (n_1\ n_2\ n_3)^\top \in \mathbb{R}^3$ is the plane normal of unit length, i.e., $|\mathbf{n}| = 1$. The signed distance of a plane $\mathbf{p}$ to a point $\mathbf{x}$ then corresponds to $\mathbf{p}\bar{\mathbf{x}}$. An allocated volume $i \in \mathbb{N}$ has its center at $\mathbf{v}_i \in \mathbb{R}^3$ and consists of $m^3$ voxels ($m = 16$ in our experiments, where each voxel has a side length of 3cm). Every volume may have up to one associated plane candidate $\mathbf{p}_i$ and a set of refined planes $\mathcal{Q}_i = \{\mathbf{q}_1, \ldots\}$. The SDF value (stored in the discrete voxel grid) for any 3D point can be retrieved using the function $\Phi : \mathbb{R}^3 \to \mathbb{R}$.

### B. Generating plane candidates

We implemented and evaluated two different ways for generating plane candidates: RANSAC and least squares.

*a) RANSAC:* By using Marching Cubes, we create a mesh from each SDF volume and run 3-point RANSAC on its vertices for finding planes. The segment is considered planar if the ratio of inliers is above a certain threshold (0.8 worked well for us). The maximum inlier distance parameter defines how aggressively planes are fitted to the surface. We found that a value of 2cm works well on Tango data and allows targeting only strictly planar surfaces. To refine the plane model, we set the normal to the smallest eigenvalue of the covariance matrix of the inlier vertices. We flip the fitted plane parameters to ensure that the normal points in the same direction as the average normal of mesh faces. Per volume, we extract at most one (the most dominant) plane candidate.

*b) Least Squares:* Instead of extracting an intermediate mesh from each SDF volume and fitting a plane to its vertices, we developed a more elegant method for directly fitting a plane to the SDF values of a volume, based on robust least squares.

We first determine the set of voxels $\mathcal{V} = \{\mathbf{x}_k\}$ inside a volume that are relevant for plane fitting, where $\mathbf{x}_k \in \mathbb{R}^3$ are the 3D coordinates of the voxel centers. Since we only want to consider voxels that contain valid distance values, we only include voxels $\mathbf{x}_k$ with $\Phi(\mathbf{x}_k) < 0.8\tau$, where $\tau$ is the SDF truncation value (10cm in our case). We also exclude voxels with weight equal to zero, i.e. unobserved regions.

To obtain the plane parameters $\mathbf{p} \in \mathbb{R}^4$ for a volume, we minimize the following least squares problem:

$$\arg\min_{\mathbf{p}} \sum_{\mathbf{x}_k \in \mathcal{V}} (r_k)^2 \quad (1)$$

with per-voxel residuals $r_k$:

$$r_k := \mathbf{p}\bar{\mathbf{x}}_k - \Phi(\mathbf{x}_k). \quad (2)$$

This formulation is equivalent to minimizing the difference between the distance from the plane to the voxel center (computed by $\mathbf{p}\bar{\mathbf{x}}_k$) and the SDF value $\Phi(\mathbf{x}_k)$ (as measurement), which encodes the distance of the voxel center to the closest surface. We determine whether the volume contains a plane based on the average of the absolute per-voxel residuals. We found that a threshold of 2cm works well to detect all local planar regions.

However, it is in practice not always the case that all voxels inside a volume actually belong to the same plane. Therefore, we integrate an additional weighting function $w$ based on the robust Huber loss function, which gives less weight to outliers, into the regular least squares formulation above. This leads to the following weighted least squares formulation:

$$\arg\min_{\mathbf{p}} \sum_{\mathbf{x}_k \in \mathcal{V}} w(r_k)(r_k)^2 \quad (3)$$

with

$$w(r) = \begin{cases} \frac{1}{2}r^2 & \text{for } |r| \leq \delta, \\ \delta(|r| - \frac{1}{2}\delta), & \text{otherwise.} \end{cases} \quad (4)$$

Empirically, we found that the value of $\delta = 5$cm works best for this task.

We minimize this problem by applying the iteratively re-weighted least squares (IRLS) algorithm, in which computation of weights and plane parameters are alternated.

### C. Merging planes

Since volumes are relatively small chunks of reconstructed geometry (in our case about half a meter in each dimension), the plane fitting described above can robustly detect the presence of a single plane within a volume. Furthermore, this allows us to detect planes independently and in parallel for every volume. This also guarantees a constant runtime since the number of updated volumes per frame is limited (due to the maximum range of the sensor), even if the total map size is growing. However, due to sensor noise, the plane estimates in adjacent volumes of the same plane are typically not identical. To reconstruct larger planar structures like walls or table tops, we therefore need to merge plane candidates at a global level and propagate them back to every affected volume. Note that this means that the number of volumes that need to be re-meshed after an update grows linearly with the map size (since all volumes sharing the same refined plane need to be updated). To keep our implementation real-time capable, we only mark those volumes for re-meshing that are within a certain radius of our current position. We implemented the following two algorithms for plane merging: RANSAC and greedy region growing.

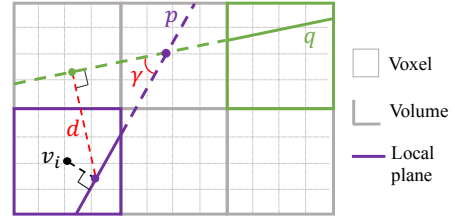

Fig. 5. Plane merging conditions. Plane $\mathbf{q}$ is merged with plane $\mathbf{p}$ if the angle $\gamma$ is below a threshold $\epsilon$ and the unsigned distance $d$ between $\mathbf{q}$ and volume center $v_i$ projected on $\mathbf{p}$ is smaller than threshold $\lambda$.

*a) RANSAC:* Our clustering approach is based on a 1-point RANSAC and adapted to find parameters of a plane that is similar to the most planes in a given set.

On each RANSAC iteration we randomly pick a plane candidate $\mathbf{p}_i$ from the set of all plane candidates in the map. We need to evaluate how many of the other plane candidates $\mathbf{p}_j$ with $i \neq j$ are inliers to $\mathbf{p}_i$. Let

$$\mathbf{x}_{ij} = \mathbf{v}_j - (\mathbf{p}_i\bar{\mathbf{v}}_j)\mathbf{n}_i, \quad (5)$$

be the projection of volume center $\mathbf{v}_j$ onto the plane $\mathbf{p}_i$. We then check the following two criteria to determine whether $\mathbf{p}_j$ is an inlier to $\mathbf{p}_i$ (see Fig. 5 for an illustration):

1) the angular difference between $q$ and $p$ is below a certain threshold (we found that 3 degrees works well for indoor environments), i.e.,

$$\mathbf{n}_i\mathbf{n}_j > \cos\epsilon \quad (6)$$

2) the Euclidean distance between $\bar{\mathbf{x}}_{ij}$ and $\mathbf{p_i}$ is below a certain threshold, i.e.,

$$|\mathbf{p_i}\bar{\mathbf{x}}_{ij}| < \lambda. \quad (7)$$

We added the latter condition to make merging of distant planes less likely even if they have similar normals ($\lambda = 5$cm in our case). After finding the largest set of inliers and refining the plane parameters, we store the refined plane for propagation (explained below). To accept a refined plane, we typically require it to have a certain number of supporting volumes, which in turn corresponds to a minimum surface area. Depending on the application, it can be useful to restrict estimation to only larger planes, like walls or large furniture surfaces.

*b) Region growing* This merging method selects one of the plane candidates in the map and executes breadth-first traversal of its neighbors, adding all volumes that have compatible plane candidates to the currently merged plane. All volumes adjacent to a volume that have similar planes are added to the traversal queue. Plane similarity is determined in the same way as in the previous approach. A drawback of this method is that outcome depends on initial plane selected for region growing.

*c) Plane propagation* In many cases SDF volumes contain more than one plane (e.g. at wall intersections, shelves or

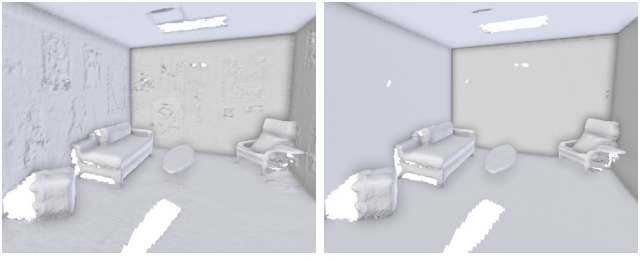

Fig. 6. Using plane priors to reduce noise in 3D reconstruction. Original reconstruction (left) and reconstruction using plane priors (right).

any small rectangular objects). To be able to query all close planes at each SDF voxel, we extend the refined planes by propagating them to all intersected volumes that are within a parametric distance. As a result, each volume can have one or more refined planes $\mathcal{Q}_i = \{q_1, \ldots\}$ associated that can be used for noise reduction, hole filling, and possibly other applications such as improved camera tracking or texturing.

### D. Noise reduction

Indoor environments are usually dominated by flat surfaces like walls, floor and shelves, but since the depth data of mobile sensors is noisy, the final 3D reconstructions contain noticeable artifacts, see Fig. 6 (left). The errors in reconstruction can also come from inaccurate localization or sparse depth data.

After we found a set of refined planes per volume as described in the previous subsection, we are now able to extract a better mesh. To this end, we define a new SDF that is based on the original SDF but replaces the distance values of voxels near a plane by the actual distance to the plane.

Let $\Phi(\mathbf{x})$ be the original SDF value for a point $\mathbf{x}$. We then define a modified SDF $\Phi'(\mathbf{x})$ based on $\Phi(\mathbf{x})$ and the set of refined planes $\mathcal{Q}_i$ of the corresponding volume $i$.

Let $\mathbf{p}_{\text{closest}}(\mathbf{x}) \in \mathcal{P}_i$ be the refined plane which is the closest to $\mathbf{x}$. Let $\mathbf{p}_{\min}(\mathbf{x}) \in \mathcal{P}_i$ be the refined plane which is has smallest signed distance to $\mathbf{x}$. Let $D_{\text{closest}}(\mathbf{x}) = \bar{\mathbf{x}} \mathbf{p}_{\text{closest}}(\mathbf{x})$ and $D_{\min}(\mathbf{x}) = \bar{\mathbf{x}} \mathbf{p}_{\min}(\mathbf{x})$ be the distances to such planes from a point $\mathbf{x}$, respectively. Then, we can compute a refined SDF value that combines the planes with existing SDF values as follows:

$$\Phi'(\mathbf{x}) = \begin{cases} D_{\min}(\mathbf{x}) & \text{if } \exists \mathbf{p}, \mathbf{q} \in \mathcal{Q}: -\tau < \bar{\mathbf{x}}^\top \mathbf{p} \cdot \bar{\mathbf{x}}^\top \mathbf{q} < 0 \\ D_{\text{closest}}(\mathbf{x}) & \text{if } |D_{\text{closest}}(\mathbf{x})| < \tau \text{ and } \\ & |D_{\text{closest}}(\mathbf{x}) - \Phi(\mathbf{x})| < \tau \\ \Phi(\mathbf{x}) & \text{otherwise} \end{cases}$$
(8)

Let us discuss every line of this equation separately:
1) The first case is targeted at voxels near plane intersections (see Fig. 7 for an illustration). We use the smallest signed distance of all close planes in areas when the refined planes are not all of the same sign. In this case, $\mathbf{x}$ is most likely located behind a plane and thus should be assigned a negative value.
2) The second case applies to voxels that are located close to a plane and thus get their distance values overwritten by the distance to this plane (see Fig. 8 for an illustration).
3) Otherwise, we take the original SDF value in areas that are far from any detected planes.

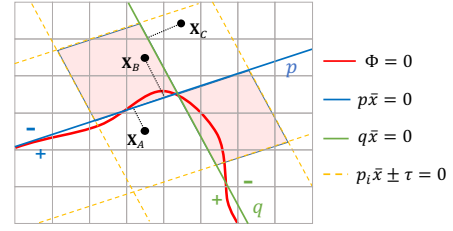

Fig. 7. Cases of SDF correction at plane intersections: at point $\mathbf{x}_A$ and $\mathbf{x}_C$ we use the distance to the nearest plane; at $\mathbf{x}_B$ and areas highlighted in red we use the lowest signed distance to the plane.

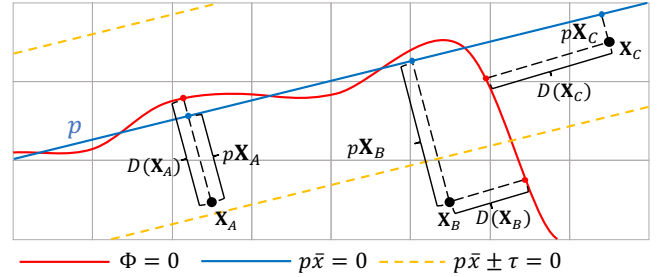

Fig. 8. Different SDF correction cases at centers of voxels $\mathbf{x}_A, \mathbf{x}_B, \mathbf{x}_C$. $\mathbf{x}_A$ - distance to plane is used since the voxel is close to the plane and not near an intersection; $\mathbf{x}_B$ - original value is used because the point is outside of truncation distance to the plane; $\mathbf{x}_C$ original value is used because difference between $D_{\text{closest}}(\mathbf{x}_C) = \mathbf{p}\bar{\mathbf{x}}_c$ and $\Phi(\mathbf{x}_C)$ is larger than truncation distance.

### E. Jitter reduction

Even after fitting planes to the model, the resulting meshing still show jitter during the online reconstruction, albeit at a lower level. To further reduce or possibly fully eliminate jitter, we only update the coefficients of those refined planes (and thus re-mesh the corresponding volumes) that differ by more than 1 degree.

### F. Hole filling

When using mobile devices for handheld 3D reconstruction, the resulting models are often incomplete, with unobserved parts of the environment appearing as holes in the reconstructed mesh. Such missing geometry also occurs behind furniture and regions occluded by objects. This can impose serious problems in many augmented reality applications, for example when objects or game characters fall through holes in the floor or enter a room through a wall. Filling holes in 3D maps is also desirable for robot navigation, for example, to prevent quadrotors to fly through apparent holes in a wall due to missing data or to enable a ground robot to plan a path due to gaps in the floor plane.

To complete the geometry where it is possible and produce smooth closed surfaces, we extend the refined planar regions and calculate SDF values in unobserved voxels from eq. 8.

This approach can also be used for generating indoor structure assumptions in the large unobserved areas. For example, the floor plane can be assumed to be infinite as long as this does not contradict any existing observations. If only distant pieces of adjacent walls have been observed, such extrapolation gives us information about their intersection even though it was not seen by the device.

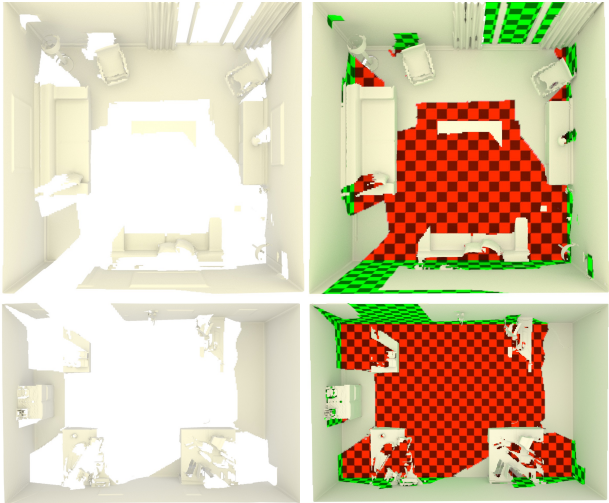

Fig. 9. Filling unobserved parts of reconstruction in 'LivingRoom-kt1' (top) and 'Office-kt1' (bottom) datasets of [20]. Our completion added 72.6% and 64.2% to the surface area respectively, recovering all walls.

Since each volume has a separate mesh in our reconstruction pipeline, we can simplify volumes which contain only planar geometry by replacing them with only two faces which form a quadrilateral. This enables more efficient storage and processing of meshes, which can be used for example in collision avoidance.

### G. Segmentation

Finding plane equations of the walls in a room is useful to cluster the scene into a set disjoint of objects.

Separation of planar regions happens naturally in both approaches of plane clustering, where we assign an ID to each global plane. Additionally, we assign one of the following semantic labels: *floor*, *wall*, *ceiling* and *other* by imposing the following rules:

1) The support area of a plane has to be greater than a threshold defined for each label. We use the number of candidates that were merged as an implicit measure of planar region area size (value of 4 worked well for floor and walls).
2) Angular difference between plane's normal and IMU gravity vector has to be within a limit set for each label. For our evaluation, we used a threshold of 10 degrees difference with unit angles.

Moreover, such labelling can be used to reconstruct only the parts of the isosurface that belong to the walls. Combined with hole filling as described above, this results in a reconstruction that contains only the room's walls, completely removing all other objects and furniture. This could be used for example in automatic 3D floor plan extraction.

TABLE I. Evaluations on the amount of added surface area by the hole filling algorithm to the mesh reconstruction with indoor datasets.

| Dataset | Raw $m^2$ | Filled $m^2$ | Improve |
|---|---|---|---|
| Poster room | 95.52 | 107.38 | 12.41% |
| Kids room | 75.66 | 95.75 | 26.55% |
| ICL-NUIM living room 1 | 64.52 | 111.35 | 72.60% |
| ICL-NUIM office 1 | 88.58 | 145.45 | 64.20% |
| BundleFusion apt0 | 113.40 | 135.77 | 19.72% |
| BundleFusion office1 | 94.40 | 137.19 | 45.33% |
| **Average** | **88.68** | **122.15** | **40.14%** |

Having all flat surfaces segmented, we can also segment furniture and other objects in the scene based on their mesh connectivity. For this, we first extract a mesh of everything that is not a plane. By traversing the mesh in breadth-first search, we can segment every vertex as part of the current object. If no unsegmented vertices can be added, the object is finalized and the process is continued to the next unsegmented vertex with a new object ID. This enables a simple but efficient decomposition of the 3D reconstruction for further object recognition.

## V. EVALUATION AND EXPERIMENTAL RESULTS

### A. Qualitative results

We evaluated our approach in ten different indoor and outdoor scenes, as shown in Fig. 10. In this experiment we used the fastest combination of methods, which is plane candidate estimation with SDF fitting and global clustering with 1-point RANSAC. We used parameter values as specified previously. All data was recorded on a Tango tablet and post-processed on a PC for visualization.

*a) Noise removal:* The second row shows the raw reconstruction from CHISEL [4] without incorporating plane priors. The walls and the floor are noisy and incomplete. The third row shows the 3D reconstruction after adding the plane prior for noise reduction. The models look much cleaner, although in certain cases small details like posters on the wall disappear.

*b) Jitter removal:* Referring back to Fig. 3, with our approach we are able to completely eliminate mesh jitter on all detected flat surfaces during online reconstruction. The video accompanying this paper (also found at https://youtu.be/8OxwiRpmzn4), we show our method used for mesh stabilization.

*c) Hole filling:* We evaluated the hole filling application by comparing the area added to the mesh using planes in unobserved regions. For this evaluation we use our datasets of rooms where all major walls are at least partially observed. We also use publicly available datasets from [20], [21], [6].

The data shows that in each scene we were able to reasonably augment the original reconstruction, adding on average 40.14% of mesh area. We conclude that our approach can recover significant amount of unobserved and occluded surfaces for indoor environments.

*d) Semantic labelling of planes:* Fig. 10 shows how our approach can be applied for detection of walls (green) and floor (red). Our system found all walls and floors with exception of kitchen dataset, where the plane fitted to wall did not receive semantic labelling due to small observed area.

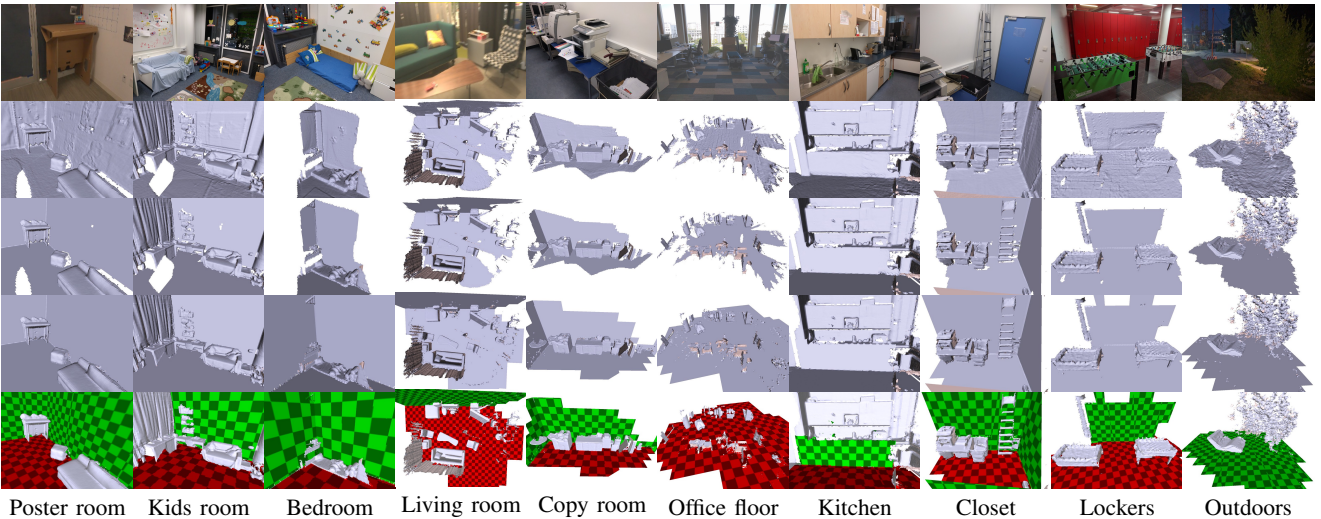

Fig. 10. Reconstruction examples. From top to bottom: original reconstruction, de-noised, holes filled, semantic segmentation of planes (green: wall, red: floor).

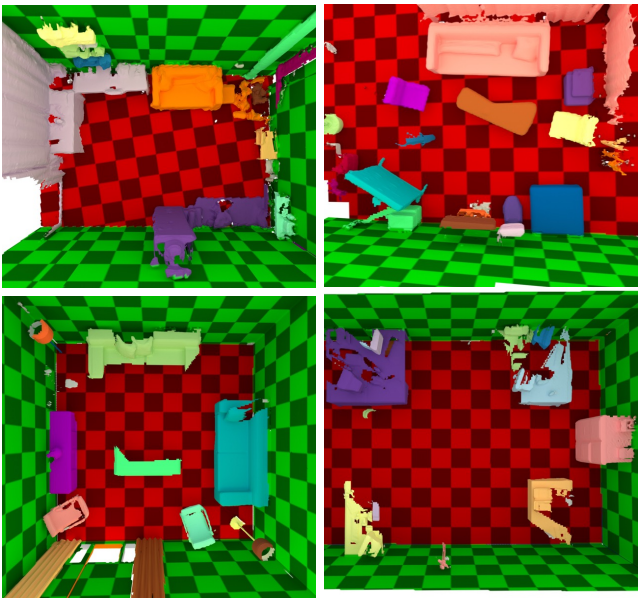

Fig. 11. Using planes for object segmentation. The planes are marked with checkerboard pattern, while separate objects are assigned different colors.

We also evaluated the number of correct classifications of walls in the scene. In these runs, 20 of 21 walls were detected correctly with the only exception of the kitchen wall that is significantly occluded by the worktop and shelves. Therefore, we conclude that our approach is capable of finding major walls our test environments at 95% accuracy.

### B. Object segmentation

Object segmentation is done as a post-processing step on a mesh with segmented planes. Fig. 11 shows that as long as there is no connectivity between objects and the walls are detected, segmentation is performed correctly. Two cases in the top row show how topologically connected objects (a curtain and a piece of furniture) are wrongly segmented as one piece.

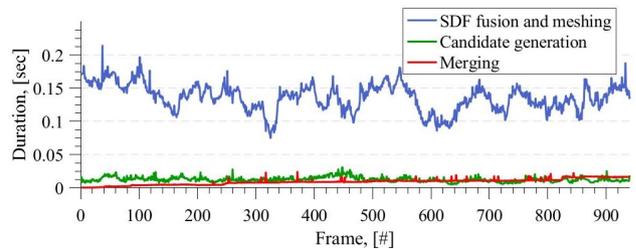

Fig. 12. Runtimes during online reconstruction of the dataset "Poster room" on a mobile device.

This method generally works well for indoor environments, where all objects are somehow attached to planar regions. On the room scale we can coarsely detect furniture and dynamic actors, whereas on the smaller scale we can segment objects laying on a table or conveyor lent that may need to be picked up with a robot arm.

### C. Runtime evaluation

We evaluated runtime performance of different methods described in this paper to find the best combination of plane fitting and merging approaches. We used a device with 4 GB RAM and 4-core processor with 2 GHz frequency for the runtime evaluation. All computational processing described in this section is performed only on the CPU. We set the maximum number of RANSAC iterations for local fitting and global association to 100 and 50 respectively. The maximum number of iterations for local fitting using iteratively re-weighted least squares on SDF is 30.

The result is given in Tab. II: The base case, where we update the SDF takes 40.8ms per frame, while extracting the mesh takes 72.8 ms. When we enable the plane prior, we need 14.5ms (or 8%) more computation time. Note that the time depends also on the scene. For example the outdoor tree generates more candidates when using RANSAC since many reconstructed auxiliary meshes are densely packed with leaves, generating more vertices resulting in more candidates and longer processing.

TABLE II. Runtime comparison of plane fitting and clustering approaches.

| Dataset | # frames | # candidate planes | | # merged planes | | Time per frame [ms] | | | | |
| --- | --- | --- | --- | --- | --- | --- | --- | --- | --- | --- |
| | | Detection method | | Merging method | | SDF fusion | Candidate generation | | Merging | |
| | | RANSAC | SDF fitting | RANSAC | Region growing | | RANSAC | SDF fitting | RANSAC | Region growing |
| Poster room | 943 | 230 | 213 | 6 | 10 | 63.2 | 85.2 | **18.9** | **3.6** | 6.8 |
| Kids room | 443 | 122 | 117 | 5 | 6 | 44.5 | 53.9 | **14.1** | **2.4** | 4.0 |
| Bedroom | 105 | 21 | 21 | 3 | 3 | 48.6 | 14.3 | **4.4** | 0.9 | **0.5** |
| Office floor | 200 | 47 | 47 | 1 | 1 | 30.5 | 33.2 | **13.0** | **0.8** | 0.9 |
| Printer | 192 | 31 | 39 | 2 | 3 | 42.3 | 25.7 | **6.8** | 1.3 | 1.4 |
| Living room | 495 | 106 | 93 | 2 | 2 | 48.4 | 62.3 | **17.9** | **2.0** | 2.5 |
| Kitchen | 224 | 32 | 32 | 3 | 3 | 30.5 | 24.3 | **11.1** | 2.3 | **2.1** |
| Closet | 325 | 51 | 51 | 4 | 4 | 44.0 | 25.5 | **7.3** | 2.0 | **1.3** |
| Lockers | 485 | 94 | 97 | 3 | 5 | 32.2 | 52.8 | **13.7** | **2.7** | 3.9 |
| Outdoors | 272 | 62 | 11 | 1 | 1 | 24.1 | 65.7 | **18.9** | 0.8 | **0.6** |
| Average | 368 | 80 | 72 | 3 | 4 | 40.8 | 44.3 | **12.6** | **1.9** | 2.4 |

While the numbers of locally found planes are similar between RANSAC and SDF fitting, the runtimes of SDF fitting is on average 3.4 times better. This is mainly caused by eliminating generation of an auxiliary mesh with Marching Cubes that is needed for RANSAC to fit a plane to vertices.

To see dynamics of processing, we measure time spent on SDF fusion, candidate generation and merging. Fig. (12) shows that processing times introduced by plane fitting are small compared to the SDF fusion, which includes depth integration and mesh generation. The time complexity of candidate generation is generally constant because the number of volumes observed in each frame is roughly the same. Plane clustering duration, however, depends linearly on the total number of candidates in the map, growing over time until the distance between the camera and some candidates reaches a pre-defined threshold.

Overall, our system is capable of online processing at interactive frame rates on a mobile device. The preferred methods for candidate generation and plane merging are least squares on SDF and 1-point RANSAC respectively.

## VI. CONCLUSION

We presented a new approach of finding and using planar surfaces during online 3D reconstruction of indoor environments. Our plane detection algorithm consists of local plane fitting and global plane merging. We show that local candidate generation can be efficiently tackled by least squares plane estimation directly on the SDF grid. Global clustering can be effectively done by 1-point RANSAC on all candidates. We demonstrated how using our approach can significantly reduce noise in online and offline 3D reconstructions and make them more visually appealing. Omitting re-meshing of planar regions resolves the vertex jitter problem of flat surfaces. By extending planar regions into unobserved and occluded parts of the scene we are able to recover missing information of room structure and fill holes already during reconstruction.